\documentclass[runningheads]{llncs}
\usepackage{amssymb}
\usepackage{amsmath}
\usepackage[T1]{fontenc}
\usepackage{graphicx,verbatim}
\usepackage{xcolor}
\begin{document}
\title{Towards 3D heart mesh generation using contactless radar imaging and physics-informed neural network }
%\titlerunning{Abbreviated paper title}
% If the paper title is too long for the running head, you can set
% an abbreviated paper title here
%
\begin{comment}  %% Removed for anonymized MICCAI submission
\author{First Author\inst{1}\orcidID{0000-1111-2222-3333} \and
Second Author\inst{2,3}\orcidID{1111-2222-3333-4444} \and
Third Author\inst{3}\orcidID{2222--3333-4444-5555}}
%
\authorrunning{F. Author et al.}
% First names are abbreviated in the running head.
% If there are more than two authors, 'et al.' is used.
%
\institute{Princeton University, Princeton NJ 08544, USA \and
Springer Heidelberg, Tiergartenstr. 17, 69121 Heidelberg, Germany
\email{lncs@springer.com}\\
\url{http://www.springer.com/gp/computer-science/lncs} \and
ABC Institute, Rupert-Karls-University Heidelberg, Heidelberg, Germany\\
\email{\{abc,lncs\}@uni-heidelberg.de}}

\end{comment}

\author{
Jinye Li\inst{1} \and
Chenxi Fu\inst{2} \and
Minghang Zheng\inst{3} \and
Yang Liu\inst{3} \and
Xiahai Zhuang\inst{2} \and
Qingchao Chen\inst{3}
}
\authorrunning{J. Li et al.}
% 运行标题：第一作者名缩写 + et al.
%
\institute{
Beijing University of Posts and Telecommunications, Beijing, China \\
\and
Fudan University, Shanghai, China \\
\and
Peking University, Beijing, China \\
\email{}
}
  
\maketitle              % typeset the header of the contribution

\begin{abstract}

Cardiac function evaluation necessitates continuous, non-invasive monitoring, a capability limited in MRI. Millimeter-wave (mmWave) radar and its Synthetic Aperture Radar (SAR) mode offer a privacy-preserving and portable point-of-care clinical applications. However, reconstructing high-fidelity 3D cardiac geometry from SAR remains an open challenge. Traditional radar methods generate sparse point clouds that lack continuous surface topology. Meanwhile, direct application of optical reconstruction networks performs poorly due to the severe speckle noise and ambiguous boundaries inherent in SAR images. To bridge this gap, we propose SAR2Mesh, a novel framework that reformulates the task as a coarse-to-fine mesh deformation process. By initializing with a topological template, our approach explicitly preserves anatomical connectivity through progressive mesh deformation.We introduce a geometry-aware feature projection module to extract multi-view features via 3D-to-2D sampling, and a physics-informed radar loss to enforce consistency between the predicted geometry and raw radar echoes. Furthermore, we present Cardiac Mesh-SAR, the first large-scale paired SAR-mesh dataset. Extensive experiments demonstrate that SAR2Mesh significantly outperforms existing image-based baselines, achieving accurate and physically consistent cardiac reconstructions.

\keywords{SAR Imaging  \and Cardiac Mesh Reconstruction \and  Graph Convolutional Networks \and Deep Learning}
% Authors must provide keywords and are not allowed to remove this Keyword section.

\end{abstract}
\section{Introduction}

\begin{figure*}[h!]
\centering
  \includegraphics[width=1\textwidth]{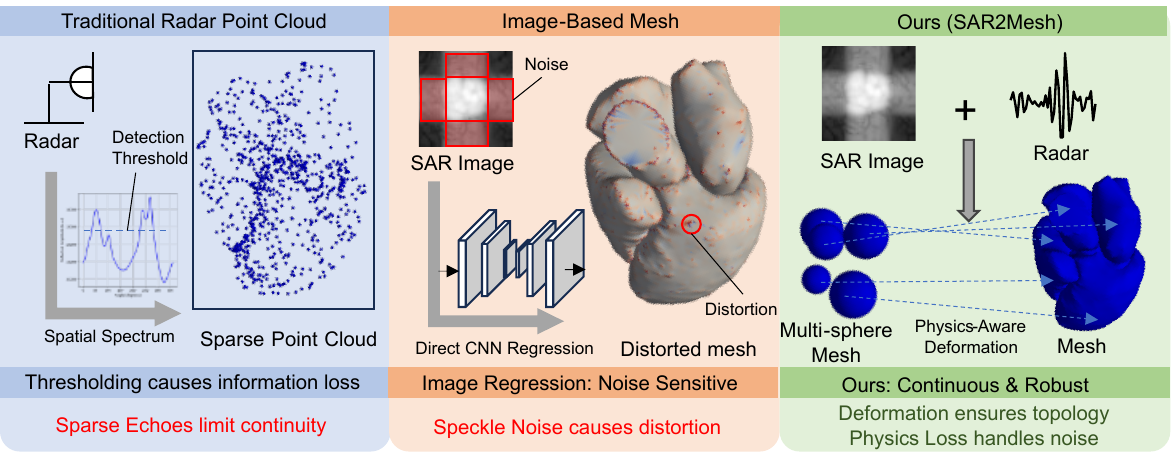}
  \caption{Comparison of cardiac reconstruction paradigms. Left: Traditional radar imaging yields sparse, fragmented point clouds. Middle: Image-based regression produces distorted meshes due to SAR speckle noise. Right: Our SAR2Mesh ensures topological continuity and robust anatomical recovery via geometry-aware deformation and physics-informed supervision.}
  \label{fig:teaser}
\end{figure*}

In clinical practice, cardiac MRI is considered the gold standard for evaluating cardiac function and morphology\cite{sun2021right}. Particularly, Cine MRI, which estimates cardiac motion, plays a crucial role in disease diagnosis and treatment planning\cite{li2020random,lu2020modelling,full2020studying}. However, due to the demanding acquisition environment and the sensitivity of the collected data, Cine MRI is not suitable for routine monitoring in everyday clinical care\cite{kong2020generalizable}.

Consequently, there is an urgent demand for a non-invasive, cost-effective, and point of care alternative that allows for continuous cardiac monitoring outside the radiology suite. In recent years, Millimeter-wave (mmWave) radar has emerged as a compelling solution, offering non-contact sensing capabilities, privacy preservation, and the ability to penetrate clothing and tissue. Specifically, Synthetic Aperture Radar (SAR) imaging holds the potential to capture high-resolution spatial information of structures. Despite these advantages, current radar applications are largely confined to 1D vital sign extraction (e.g., heart rate) or superficial body shape estimation.

Reconstructing detailed 3D cardiac geometry directly from SAR signals remains an unexplored frontier. Existing approaches fall short in this domain. Specifically, traditional radar signal processing techniques \cite{schmidt1986multiple,roy2002esprit} excel at point source localization via subspace analysis. However, as illustrated in Figure \ref{fig:teaser}, these methods fail to preserve the continuous surface topology and complex anatomical connectivity required for organ modeling due to the sparse and discrete nature of radar returns. Such topological continuity is indispensable for accurate ventricular volume calculation and functional assessment in clinical diagnosis. Addressing this limitation requires explicit surface modeling, where graph-based deformation networks \cite{wen2019pixel2mesh++} have demonstrated efficacy in iteratively refining coarse templates. Nevertheless, directly adapting such image-based reconstruction networks to the SAR domain is ineffective. While recent computer vision paradigms have shifted towards implicit neural representations \cite{mildenhall2021nerf,muller2022instant} and surface-focused variants \cite{yariv2021volume,wang2021neus,li2023neuralangelo} for photorealistic synthesis, their application to the sparse, speckle-dominated nature of medical radar data remains non-trivial. Unlike optical images, SAR images are dominated by speckle noise and lack clear boundaries, rendering standard feature extraction techniques insufficient for capturing fine geometric details.

To address these limitations, we present SAR2Mesh, a novel framework designed to robustly reconstruct high-fidelity 3D cardiac meshes from multi-view SAR images. Unlike point-based approaches, we adopt a deformation-based paradigm using Graph Convolutional Networks (GCNs). By initializing with a template mesh and leveraging its inherent topological priors, our model ensures surface continuity and anatomical plausibility, effectively overcoming the fragmentation issues of point cloud methods. Furthermore, to tackle the ambiguity of SAR features, we propose a coarse-to-fine deformation strategy coupled with a geometry-aware multi-view feature projection module. This allows the network to progressively resolve geometric details—from global shape to local refinements—by aggregating multi-scale contextual information, thereby mitigating the impact of speckle noise and unclear boundaries inherent in radar imaging.

Beyond the methodological challenges, the advancement of this field is further hindered by the absence of high-quality paired data. To bridge this gap and establish a rigorous benchmark for our framework, we introduce Cardiac Mesh-SAR , the first large-scale hybrid dataset containing paired multi-view SAR images and 3D cardiac meshes. Cardiac Mesh-SAR is designed to evaluate reconstruction across varying anatomical complexities: it includes a whole-heart subset derived from Statistical Shape Models (SSM) for validating complex global topology reconstruction, and a biventricular subset from public and in-house clinical MRI data for assessing performance on dynamic cardiac structures.

In summary, our main contributions are as follows:

\textbf{We propose SAR2Mesh, a pioneering deep learning framework that reconstructs high-fidelity 3D cardiac meshes directly from mmWave SAR signals.} By employing a coarse-to-fine graph convolutional network with geometry-aware multi-view feature projection, our method effectively hallucinates missing geometric details from sparse radar returns, enabling non-invasive internal organ reconstruction.

\textbf{We incorporate a novel physics-informed differentiable radar loss to regularize the reconstruction process.} This mechanism enforces consistency between the predicted geometry and the raw radar signal in the simulation domain, effectively constraining the solution space to mitigate speckle noise and resolve geometric ambiguities inherent in radar imaging.

\textbf{We introduce Cardiac Mesh-SAR, the first large-scale benchmark dataset for this task, comprising paired multi-view SAR images and 3D meshes.} By integrating SSM-augmented whole-heart data and clinically relevant biventricular data, Cardiac Mesh-SAR provides a comprehensive testbed for future research in radar-based medical imaging.

\section{Method}

\begin{figure*}[h!]
\centering
  \includegraphics[width=1\textwidth]{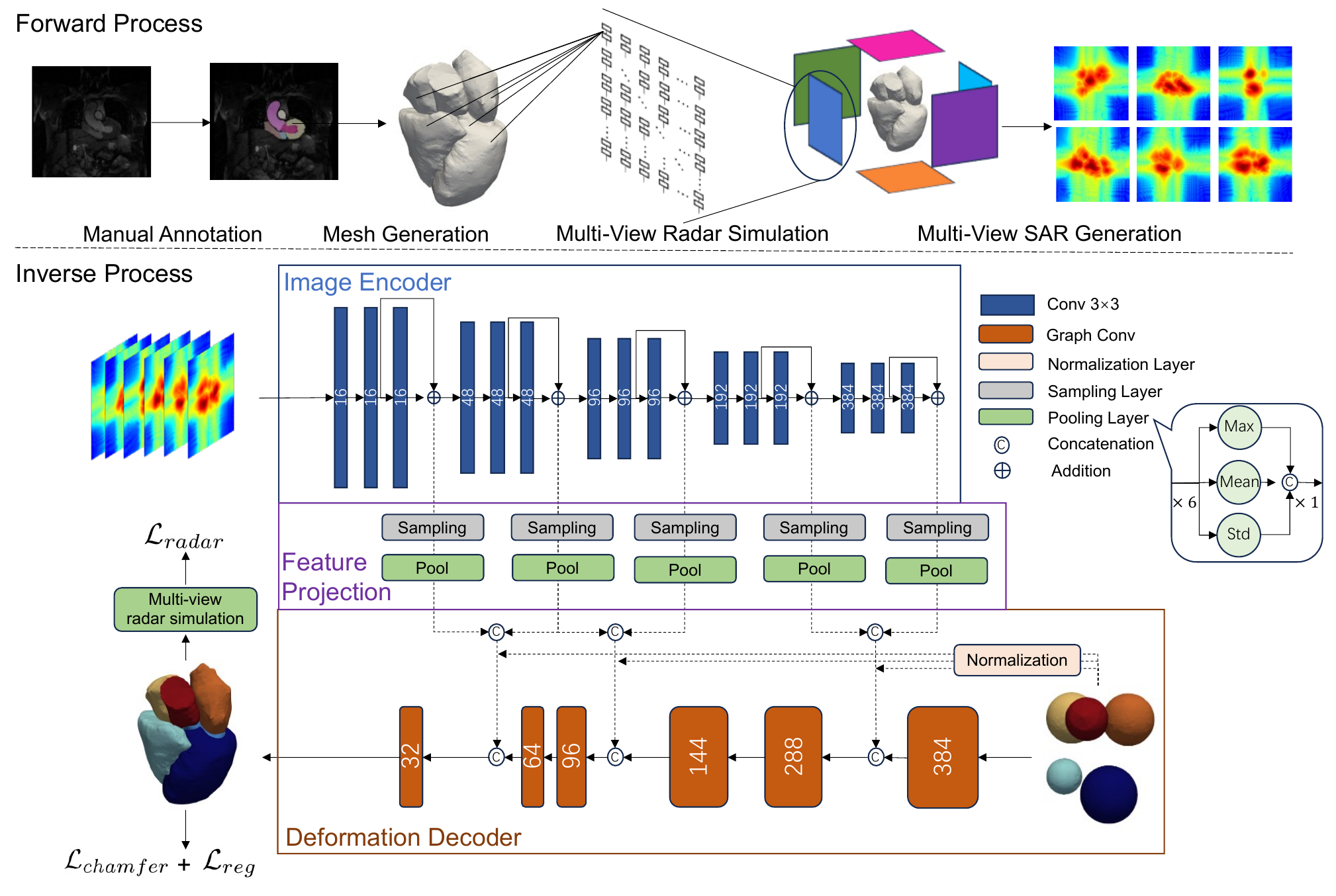}
  \caption{Overview of SAR2Mesh. Top: Forward Simulation generates paired meshes and multi-view SAR images from MRI. Bottom: Inverse Reconstruction employs a Shared Image Encoder to extract multi-scale features. A Feature Projection module dynamically samples and pools these features based on 3D vertex positions, guiding a Deformation Decoder to progressively deform a template sphere into a high-fidelity cardiac mesh under geometric and physics-informed supervision.}
  \label{method}
\end{figure*}

\subsection{Overview}
Reconstructing 3D cardiac geometry from sparse, speckle-dominated mmWave radar signals is an ill-posed inverse problem where traditional point-based methods fail to preserve topology and image-based regressors struggle with noise. SAR2Mesh addresses this by reformulating reconstruction as a coarse-to-fine mesh deformation task. Our framework comprises: (1) a \textbf{Physics-Based Forward Simulation} to synthesize paired data; (2) a \textbf{Geometry-Aware Inverse Reconstruction} network that deforms a template mesh using multi-view features; and (3) a \textbf{Physics-Informed Radar Loss} to enforce signal-domain consistency.

\subsection{Forward Process: Physics-Based SAR Simulation}
\label{sec:method_forward}
To bridge the supervision gap, we synthesize paired radar-mesh data. From MRI-derived meshes, we model the mmWave IF signal $s(t)$ as a superposition of reflections from visible facets $f_k$:
\begin{equation}
s(t) = \sum_{k} \mathbb{I}(v_k) \cdot \mathcal{A}(f_k, \theta_k) \cdot e^{j \Phi(t, \tau_k)}
\end{equation}
where $\mathbb{I}(v_k)$ is the visibility indicator, $\mathcal{A}$ accounts for specular reflection and geometric spreading, and $\Phi$ encodes the FMCW phase. We then employ a matched-filtering based wavenumber algorithm\cite{chen2017coherent} to reconstruct 3D spatial reflectivity, providing the input for our inverse network.

\subsection{Inverse Problem: SAR-to-Mesh Reconstruction}
\label{sec:method_inverse}
Unlike point-cloud methods, we guarantee topological continuity by evolving a spherical template $\mathcal{M}^{(0)}$. Our pipeline tackles radar ambiguity through three key components.

\noindent \textbf{Multi-scale Image Encoding.} To robustly capture both global shape and local details from speckle-dominated SAR images, we employ a shared 2D encoder $E_\theta$ based on a UNet architecture. Given $V$ orthogonal views, each image $I_v$ is processed independently to produce a hierarchy of feature maps $\mathcal{F}_v = \{F_v^{(l)} \mid l=0, \dots, L-1\}$.
Sharing weights across views ensures view-invariant feature learning, which is critical for consistent geometric reasoning. The multi-scale design allows the network to utilize deep, low-resolution semantic features to recover the global shape structure during initial deformation, while leveraging shallow, high-resolution features to capture fine-grained local details in subsequent refinement stages.

\noindent \textbf{Geometry-Aware Feature Projection.} Standard CNNs lack explicit 3D-to-2D correspondence. We bridge this gap by projecting 3D vertices onto 2D image planes to sample features. For a vertex $p_i$, its projected coordinate in view $v$ is $u_{i,v} = \Pi_v(p_i)$. We sample features $f_{i,v}^{(l)}$ via bilinear interpolation:
\begin{equation}
    f_{i,v}^{(l)} = \text{GridSample}(F_v^{(l)}, u_{i,v})
\end{equation}
A critical challenge in radar imaging is the severe occlusion and signal dropout inherent in single views, where anatomical regions are often only partially visible or entirely missing due to limited penetration and specular reflection. To obtain a robust vertex descriptor, we aggregate features from all $V$ views using a combination of max-pooling (to capture the strongest signal), average-pooling (for smoothing), and standard deviation pooling (to encode signal variability):
\begin{equation}
    \tilde{f}_i^{(l)} = \text{Concat}\left( \max_{v} f_{i,v}^{(l)}, \frac{1}{V}\sum_{v} f_{i,v}^{(l)}, \text{Std}_{v} f_{i,v}^{(l)} \right)
\end{equation}
This aggregated feature $\tilde{f}_i^{(l)}$ effectively mitigates the sparsity issue, providing a view-invariant representation for geometry reasoning.

\noindent \textbf{Coarse-to-Fine Deformation Decoder.} Direct regression of high-frequency details from noisy radar signals is prone to local minima. We adopt a coarse-to-fine strategy using a cascaded Graph Convolutional Network (GCN) decoder. The deformation starts from a spherical template $\mathcal{M}^{(0)}$ and proceeds through $K=3$ stages.
At each stage $k$, the network predicts a vertex displacement $\Delta p_i^{(k)}$ using a specific subset of feature scales $\mathcal{S}_k$. Early stages use deep features to recover the global hull, while later stages use shallow features to refine local boundaries. The update rule is:

\begin{align}
    h_i^{(k)} &= \text{Concat}\left( p_i^{(k-1)}, \phi^{(k-1)}_i, [\tilde{f}_i^{(l)}]_{l \in \mathcal{S}_k} \right) \\
    p_i^{(k)} &= p_i^{(k-1)} + \text{GCNBlock}_k(h_i^{(k)}, \mathcal{A})
\end{align}

where $\text{GCNBlock}$ consists of a graph convolution layer followed by batch normalization and ReLU. $\phi^{(k)}_i$ is the hidden geometric feature propagated between stages. This progressive refinement ensures topological stability while maximizing detail recovery.

\subsection{Optimization}
\label{sec:method_loss}
Our composite loss function $\mathcal{L}_{total}$ combines geometric supervision with a novel physics constraint:
\begin{equation}
    \mathcal{L}_{total} = \lambda_{chamf} \mathcal{L}_{chamfer} + \lambda_{reg} \mathcal{L}_{reg} + \lambda_{radar} \mathcal{L}_{radar}
\end{equation}

\noindent \textbf{Geometric Regularization.} We use Chamfer Distance ($\mathcal{L}_{chamfer}$) for surface accuracy, regularized by Laplacian smoothness, edge length, and normal consistency terms ($\mathcal{L}_{reg}$) to prevent artifacts.

\noindent \textbf{Physics-Informed Radar Loss.} To capture radar-specific characteristics like multipath, we introduce a differentiable radar simulation module. We minimize the $L_1$ distance between the simulated signal of the predicted mesh and the ground truth signal:
\begin{equation}
    \mathcal{L}_{radar} = \| s_{pred}(t) - s_{gt}(t) \|_1
\end{equation}
where $s_{pred}(t)$ is computed via differentiable soft visibility. We calculate the probability of visibility using a sigmoid function applied to the angle between the surface normal and view direction, providing smooth gradients for shape deformation. This enforces the reconstructed anatomy to be physically consistent with the observed echoes.

\section{Experiment}

\subsection{Experiment Setup}

\noindent \textbf{Datasets \& Metrics:} We evaluate SAR2Mesh on three Cardiac Mesh-SAR datasets, where SAR data is synthesized from MRI-derived meshes via our forward simulation: (1) \textbf{MMWHS-SSM} (1,000 meshes generated by Statistical Shape Modeling\cite{unberath2015open} based on MMWHS\cite{zhuang2016multi}), (2) \textbf{ACDC}\cite{bernard2018deep}, and (3) \textbf{Private Dataset} (in-house patient MRI). The dynamic meshes for (2) and (3) are generated from masks predicted for each frame based on manual annotations\cite{ding2025cinemyops}. For all datasets, we use 80\% of the data for training and 20\% for testing. Reconstruction quality is assessed using seven metrics: volumetric overlap (Dice, Jaccard), surface accuracy (ASSD, HD), and geometric properties (ANE, ANLD, SI\cite{htetgen201541211delaunay}).

\noindent \textbf{Baselines \& Implementation:} We exclude traditional radar point estimation methods as they produce sparse point clouds incomparable to mesh metrics. Instead, we benchmark against \textbf{Pixel2Mesh++}\cite{wen2019pixel2mesh++}, \textbf{NeuS}\cite{wang2021neus}, \textbf{VolSDF}\cite{yariv2021volume}, and \textbf{Neuralangelo}\cite{li2023neuralangelo}, all retrained on our data. Our model is trained for 500 epochs (Adam optimizer, lr=$1e^{-3}$, batch size=1) with a radar loss weight of 0.1 (20-epoch warmup) and stage-wise geometric weights $[0.3, 0.05, 0.46, 0.16]$.

\begin{table*}[th!] 
  \centering  
    \caption{Comparison of results with other models on the MMWHS-SSM Cardiac Mesh-SAR dataset.} 
    \label{SSM} 
    \resizebox{\textwidth}{!}{ 
    \begin{tabular}{c|ccccccc} 
    \hline 
      & Dice$\uparrow$ & Jaccard$\uparrow$ & ASSD$\downarrow$ & HD$\downarrow$ & ANE$\downarrow$  & ANLD$\downarrow$ & Intersection$\downarrow$ \\ 
     \hline 
    Pixel2Mesh++      & 0.939 & 0.885 & 0.0416 & 0.0902 & 0.233 & 0.532 & 29.695\\ 
      NeuS            & 0.966 & 0.934 & 0.0272 & 0.119 & 0.176 & 0.409 & 23.574\\ 
      VolSDF          & 0.954 & 0.912 & 0.0285 & 0.121 & 0.181 & 0.425 & 24.110\\ 
      Neuralangelo    & 0.981 & 0.963 & \textbf{0.0219} & 0.0761 & 0.146 & 0.322 & \textbf{20.959}\\ 
      Ours            & \textbf{0.992} & \textbf{0.983} & 0.0223 & \textbf{0.0754} & \textbf{0.135} & \textbf{0.309} & 21.913\\ 
      \hline 
    \end{tabular} 
    } 
\end{table*}
\begin{table*}[th!] 
  \centering  
    \caption{Comparison of results with other models on the ACDC Cardiac Mesh-SAR dataset.} 
    \label{acdc} 
    \resizebox{\textwidth}{!}{ 
    \begin{tabular}{c|ccccccc} 
    \hline 
      & Dice$\uparrow$ & Jaccard$\uparrow$ & ASSD$\downarrow$ & HD$\downarrow$ & ANE$\downarrow$  & ANLD$\downarrow$ & Intersection$\downarrow$ \\ 
     \hline 
    Pixel2Mesh++           & 0.709 & 0.549 & 0.0693 & 0.185 & 1.315 & 0.291 & 36.916\\ 
      NeuS                 & 0.772 & 0.628 & 0.0489 & 0.247 & 1.136 & 0.183 & 31.259\\ 
      VolSDF               & 0.765 & 0.620 & 0.0514 & 0.263 & 1.182 & 0.199 & 31.993\\ 
      Neuralangelo         & 0.796 & 0.661 & 0.0402 & 0.159 & \textbf{0.997} & 0.144 & \textbf{27.975}\\ 
      Ours                 & \textbf{0.801} & \textbf{0.668} & \textbf{0.0386} & \textbf{0.142} & 1.010 & \textbf{0.126} & 29.827\\ 
      \hline 
    \end{tabular} 
    } 
\end{table*}
\begin{table*}[th!] 
  \centering  
    \caption{Comparison of results with other models on the private Cardiac Mesh-SAR dataset.} 
    \label{private} 
    \resizebox{\textwidth}{!}{ 
    \begin{tabular}{c|ccccccc} 
    \hline 
      & Dice$\uparrow$ & Jaccard$\uparrow$ & ASSD$\downarrow$ & HD$\downarrow$ & ANE$\downarrow$  & ANLD$\downarrow$ & Intersection$\downarrow$ \\ 
     \hline 
    Pixel2Mesh++           & 0.795 & 0.660 & 0.198 & 0.281 & 0.204 & 0.267 & 32.513\\ 
      NeuS                 & 0.858 & 0.752 & 0.143 & 0.316 & 0.139 & 0.162 & 27.896\\ 
      VolSDF               & 0.847 & 0.735 & 0.159 & 0.323 & 0.151 & 0.175 & 28.971\\ 
      Neuralangelo         & 0.889 & 0.801 & 0.114 & 0.238 & \textbf{0.106} & 0.129 & \textbf{23.652}\\ 
      Ours                 & \textbf{0.903} & \textbf{0.824} & \textbf{0.100} & \textbf{0.226} & 0.110 & \textbf{0.102} & 24.929\\ 
      \hline 
    \end{tabular} 
    } 
\end{table*}

\subsection{Results}

Quantitative results in Table \ref{SSM}, Table \ref{acdc}, and Table \ref{private} demonstrate that SAR2Mesh consistently outperforms baselines across most metrics. On the MMWHS-SSM Cardiac Mesh-SAR dataset, our method achieves the highest volumetric overlap and surface accuracy, validating our coarse-to-fine deformation strategy for global topology capture. Although implicit methods like Neuralangelo show competitive self-intersection rates, our approach maintains superior geometric fidelity.

On the challenging dynamic ACDC and Private Cardiac Mesh-SAR datasets, SAR2Mesh maintains its lead in volumetric and surface metrics, surpassing the second-best method, Neuralangelo, by a notable margin. While implicit representations inherently minimize self-intersections, our explicit deformation yields significantly better geometric accuracy (lower HD and ASSD). This advantage stems from our \textbf{physics-informed} paradigm, which integrates differentiable radar simulation into the optimization loop. By enforcing signal-domain consistency, our method effectively regularizes geometry against speckle noise, enabling precise boundary reconstruction that purely data-driven implicit fields struggle to achieve.

\subsection{Qualitative Analysis}

\begin{figure*}[t!]
\centering
  \includegraphics[width=1\textwidth]{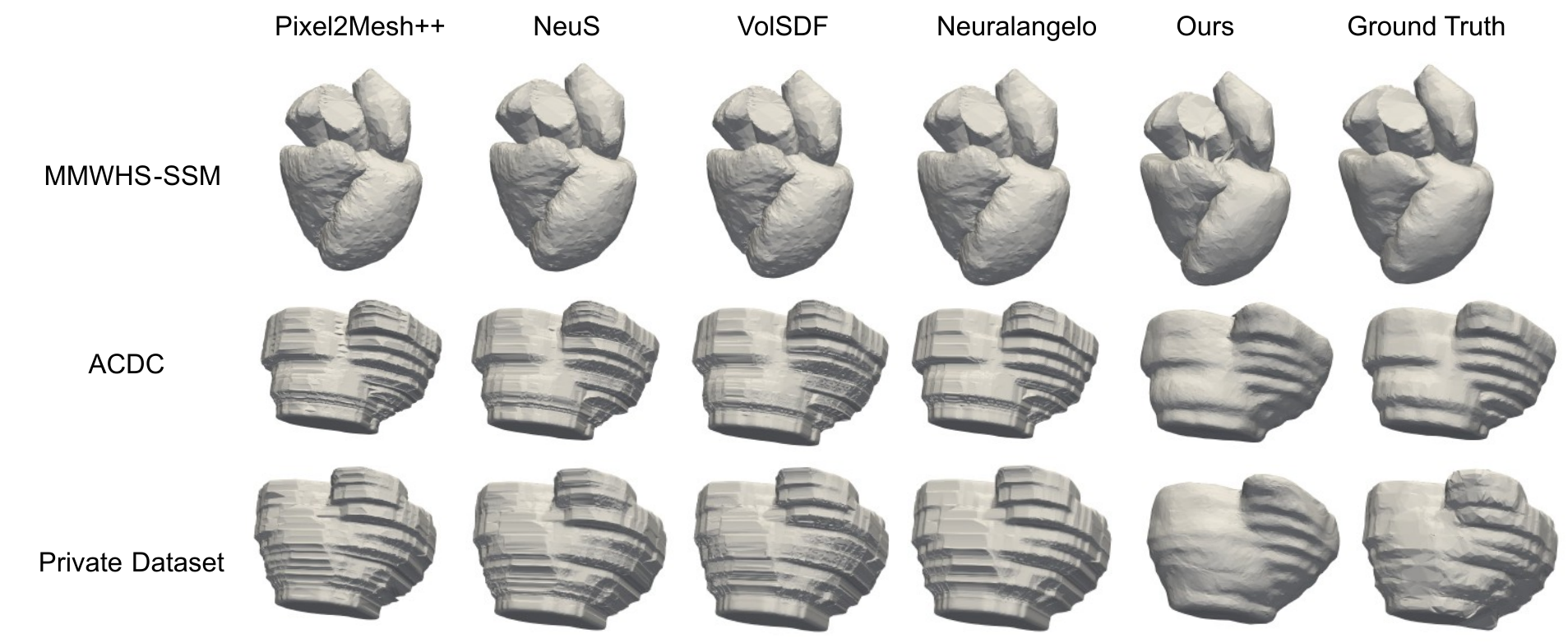}
  \caption{Qualitative comparison of 3D reconstruction results across MMWHS-SSM (top), ACDC (middle), and Private Dataset (bottom). We compare our method baselines: Pixel2Mesh++, NeuS, VolSDF, and Neuralangelo. }
  \label{fig:exp}
\end{figure*}

We conduct a comprehensive visual comparison of the reconstructed 3D meshes across the MMWHS-SSM, ACDC, and Private Cardiac Mesh-SAR datasets. As shown in Figure \ref{fig:exp}, baseline methods including Pixel2Mesh++, NeuS, VolSDF, and Neuralangelo struggle to handle the inherent speckle noise and sparsity of radar signals. Their reconstructed surfaces exhibit significant high-frequency artifacts, manifesting as irregular spikes and noise, particularly in regions that require smooth anatomical transitions. In contrast, SAR2Mesh generates clean, topologically continuous surfaces that closely align with the Ground Truth. Our method effectively suppresses surface noise while faithfully preserving critical structural details, avoiding the excessive roughness seen in baseline methods.

% \textcolor{blue}{@JINYE, don't forget to qualitatively visualize some results where different methods have different reconstruction errors; showcase when and where in 3d meshin details our method outperforms the other methods. }

\section{Conclusion}

We proposed SAR2Mesh, a novel framework for reconstructing high-fidelity 3D cardiac geometry from mmWave SAR images. By integrating geometry-aware mesh deformation with a physics-informed radar loss, our method robustly recovers anatomical details, overcoming the limitations of traditional point-based and image-regression approaches. We also introduced Cardiac Mesh-SAR, the first large-scale paired benchmark for this domain. Extensive experiments demonstrate that SAR2Mesh significantly outperforms state-of-the-art baselines, validating the potential of mmWave radar as a non-invasive modality for continuous cardiac monitoring. 

\bibliographystyle{IEEEtran}
\bibliography{IEEEabrv,refs}

\end{document}